\documentclass[10pt,twocolumn,letterpaper]{article}

\usepackage[]{cvpr}      
\usepackage{multirow}
\usepackage{float}
\usepackage{makecell}
\usepackage[utf8]{inputenc}   
\usepackage{csquotes}         
\usepackage{svg}
\usepackage{cuted} 
\usepackage{caption}
\usepackage{balance}  
\usepackage{graphicx}
\usepackage{stfloats}
\usepackage{multicol}

\usepackage{pdftexcmds} 
\usepackage[final]{changes} 
\definechangesauthor[name={dzy}, color=orange]{dzy}
\definechangesauthor[name={dzy0515}, color=orange]{dzy0515}

\definecolor{cvprblue}{rgb}{0.21,0.49,0.74}
\usepackage[pagebackref,breaklinks,colorlinks,allcolors=cvprblue]{hyperref}

\def\paperID{} 
\def\confName{CVPR}
\def\confYear{2026}

\title{TinySAM 2: Extreme Memory Compression for Efficient Track Anything Model}

\author{Zhaoyuan Ding, Yijing Yang, Han Shu, Xinghao Chen\\
Huawei Noah’s Ark Lab\\
{\tt\small  \{dingzhaoyuan2, yangyijing4, han.shu, xinghao.chen\}@huawei.com}
}

\begin{document}
\maketitle

\begin{abstract}

Segment Anything Model 2 (SAM 2) serves as a core foundation model in the field of video segmentation. Building upon the original SAM model, it introduces a memory bank mechanism and demonstrates outstanding performance in tasks such as semi-supervised video object segmentation and tracking anything. However, the complex computational characteristics of SAM 2's multi-stage image encoder and memory module have raised the barrier to the model's deployment in practical applications. To address this issue, we propose TinySAM 2, a lightweight video segmentation model that balances performance and efficiency. First, a memory quality management mechanism is introduced to select and retain high-informative historical frames as the memory. In addition, a joint-spatial-temporal token compression is proposed that reduces the memory storage and computational cost. Specifically, average pooling is employed to first compress redundancy tokens in the spatial domain. In the temporal domain, informative tokens are selected across frames in the memory bank based on token-level similarity measurement. Besides, we take RepViT as the lightweight image encoder, which further reduces the model parameters. Extensive experiments on challenging datasets such as DAVIS and SA-V demonstrate that TinySAM 2 achieves 90\% of the performance of SAM 2.1, with only 7\% memory tokens and 3\% training data. This study effectively alleviates the bottlenecks in parameter count, computational load, and deployment costs associated with SAM 2, providing a resource-efficient solution for the widespread application of video segmentation models on devices.

\end{abstract}

\section{Introduction } \label{sec:intro}

\begin{figure}[!ht]
    
    \includegraphics[height=5.5cm]{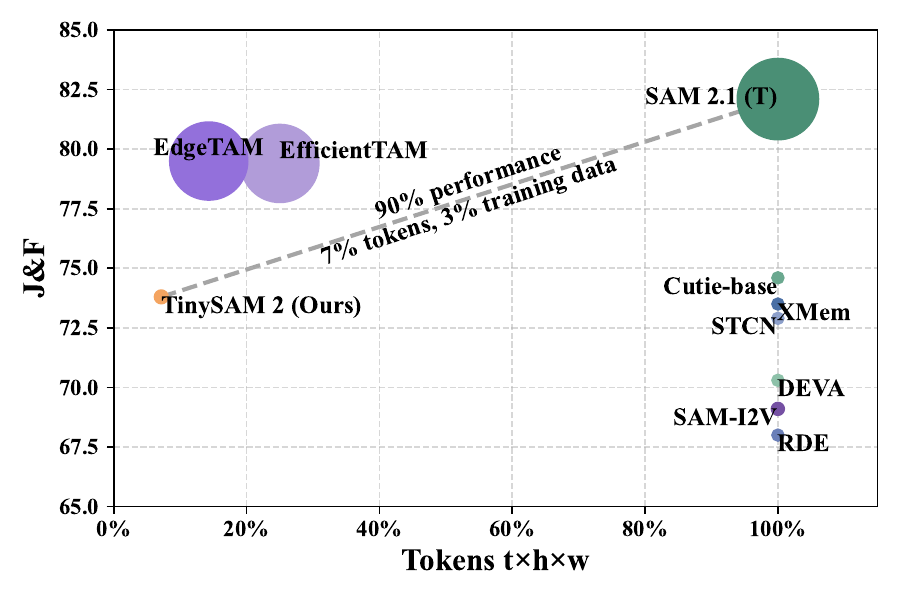}
    \caption{\textbf{Standard Semi-Supervised Video Object Segmentation Benchmark Comparison.} The average $\mathcal{J\& F}$ performance (y-axis) of TinySAM 2, SAM 2.1, SAM-I2V and other models on datasets such as SA-V, plotted against token length (x-axis). The size of the circles represents the amount of data used for model training.\added[id=dzy]{ These data are from Table~\ref{tab:c4-1-compare-to-SOTA}, , and the values for $\mathcal{J\& F}$ are the averages.} \replaced[id=dzy]{Except for TinySAM 2 and SAM-I2V, all other models have a pre-training phase. SAM 2.1, EdgeTAM, and EfficientTAM use the full SA-1B dataset during the pre-training phase. In addition, SAM 2.1 also uses an undisclosed internal dataset, so their data volume is significantly larger than that of the other models.}{Since SAM 2.1, EdgeTAM, and EfficientTAM use the full SA-1B dataset during the pretraining phase, their data volume is significantly larger than that of other models.} TinySAM 2 achieves 90\% of the performance of SAM 2.1 despite having the shortest token length and the much less training data.}
	\vspace{-.35cm}
    \label{fig_c2_compare}
\end{figure}

Segment Anything Model (SAM)~\cite{Kirillov_2023_ICCV}  is the first promptable foundational model for image segmentation, establishing a foundational framework for image-level segmentation. Its successor, SAM 2~\cite{ravi2024sam2segmentimages}, further advances this paradigm toward unified spatiotemporal modeling: it extracts hierarchical frame features through a multi-granularity image encoder and continuously integrates historical appearance and positional priors using a cross-frame memory attention mechanism, achieving state-of-the-art (SOTA) performance in zero-shot image segmentation, semi-supervised video object segmentation, and interactive tracking. Trained on the large-scale video dataset SA-V~\cite{ravi2024sam2segmentimages}, SAM 2 significantly improves long-term target consistency in complex scenes, opening new avenues for the transfer and generalization of foundational models in the dynamic vision domain.

However, SAM 2's exceptional accuracy comes at a high computational cost, which becomes a critical bottleneck for deployment on devices. {Its default HieraB+~\cite{pmlr-v202-ryali23a} encoder has around 80M parameters which is too heavy for edge computation. More importantly, }
the memory attention module introduced in SAM 2 involves cross-attention computation between the current frame features and the memory frame features. In the matrix multiplication computation, the token length involved in the memory frame features remains around 30,000~\cite{xiong2024efficienttrack}. As a result, although the memory attention module has relatively few parameters, its computational complexity is still too high for device-side inference. More critically, training SAM 2 from scratch requires 256 A100 GPUs with 80GB memory each for 108 hours, creating a significant barrier to exploring efficient architectures due to the enormous computational requirements. Therefore, how to compress the image encoder and memory attention module while maintaining prompt-driven spatiotemporal segmentation capabilities, and to reduce training resources, has become a central issue in advancing the adoption of prompt-based video segmentation foundational models.

Based on the above observations, {we propose a lightweight and efficient solution that equipped with two strategies: memory quality management and memory token compression, to balance the segmentation quality and computational cost, laying a feasible foundation for device-side deployment at the architectural level. We propose a spatiotemporal compression that can effectively select the most informative tokens for segmentation and tracking at an extreme compression ratio. In the spatial dimension, we compress intra-frame features through spatial pooling operations before pushing an incoming frame into the memory bank. In the temporal dimension, we reduce the redundancy of features between adjacent frames through a inter-frame feature comparison mechanism. Only the tokens with lower similarity across frames are selected since they carry the most motion information which is critical to video segmentation and tracking. By trimming redundant tokens between frames, we significantly reduce redundant computation and further optimize the model's inference efficiency. As for the memory quality management, we use a mixture of measurements based on IOU metric and absence filtering. The former is used to quantify the correlation between historical memory and current frame features, while the latter is used to remove invalid memory information.
These two strategies work together to accurately select and retain high-quality memories, significantly reducing memory storage costs while maintaining the effectiveness of feature representation. Besides, we use RepViT~\cite{Wang_2024_CVPR} as the lightweight image encoder to further reduce the model parameters.}

To verify the effectiveness of our method, we conducted comprehensive experiments on challenging datasets such as SA-V~\cite{ravi2024sam2segmentimages} and DAVIS~\cite{ponttuset20182017davischallengevideo}. The results show that our model requires only $7\%$ of the token length, $3\%$ of the training data, and $3.8\%$ of the training resources compared to SAM 2.1~\cite{ravi2024sam2segmentimages} as illustrated in Figure~\ref{fig_c2_compare}; yet, {the performance is comparable which reaches at $90\%$ of that of SAM 2.1.} These optimizations not only achieve segmentation and tracking performance on par with SOTA models, but also significantly lower the training thresholds of SAM 2.1 with a real-time inference speed, successfully overcoming the constraints of device-side hardware resources. 

{To this ends, we propose an efficient video object segmentation and tracking framework, TinySAM 2, which is designed for device-side deployment.} Our contributions are summarized as follows:
\begin{itemize}
\item For the first time, we jointly optimize memory management and joint-spatial-temporal compression mechanisms based on a lightweight image encoder, which significantly reduces the computational and memory overhead that can run in real-time.

\item In the spatial {domain}, we introduce spatial pooling operations to effectively compress intra-frame features, and further integrate Intersection over Union(IOU) metrics with an absence filtering strategy. 
This approach greatly reduces memory storage and computational costs while maintaining the effectiveness of feature representation.

\item In the temporal {domain, we propose to calculate the similarity of features between adjacent frames which reduces inter-frame redundant tokens. It significantly decreases the cost in the memory attention computation, yet maintains a high accuracy in segmentation and tracking tasks.}
\end{itemize}

To the best of our knowledge, this is the first device-side video object segmentation model 
that jointly incorporates {memory management and spatiotemporal memory token compression}. It balances efficiency and accuracy, opening up new directions for the application of lightweight visual foundational models in mobile scenarios.
\section{Related Work} \label{sec:related_work}

\noindent\textbf{Video Object Segmentation (VOS). }\deleted[id=dzy0515]{VOS  aims to perform pixel-level tracking and segmentation on subsequent frames given an initial frame with a real object mask~\cite{ponttuset20182017davischallengevideo}. The input mask can be considered as the only supervisory signal for the object in the video, hence this task is referred to as \enquote{semi-supervised video object segmentation} (SVOS)~\cite{caelles2017one,wang2018semi,oh2019video,bhat2020learning,park2021learning,fan2021semi,sun2023munet}. 
Early {work} relies on recurrent neural networks for spatiotemporal encoding~\cite{Li_2022_CVPR}, while recent Transformer-based models~\cite{Cheng2023PuttingTO, 10943417, 10487964} have achieved significant breakthroughs in accuracy and robustness due to their global self-attention and long-range dependency modeling capabilities. }
\added[id=dzy0515]{Recent Transformer-based models~\cite{Cheng2023PuttingTO, 10943417, 10487964} have achieved significant breakthroughs}
VOS that can segment salient objects without any annotations is known as unsupervised Video Object Segmentation (UVOS)~\cite{hu2018unsupervised,li2018unsupervised,lu2019see,wang2019learning}, {while} interactive VOS (IVOS) iteratively refines the mask through lightweight cues such as clicks or bounding boxes~\cite{cheng2021modular,miao2020memory,oh2019fast,heo2020interactive}. 
SAM 2 further unifies SVOS and IVOS into Promptable Visual Segmentation (PVS), which allows flexible input of multiple prompts at any frame. This significantly lowers the barrier for obtaining high-quality masks in the first frame, providing more efficient and flexible solutions for practical applications such as video editing and robotics.

\noindent\textbf{Segment Anything Model (SAM). }SAM~\cite{Kirillov_2023_ICCV} defines a new prompt-based segmentation task where users can provide prompts through points, boxes, and masks. {Several follow-up work aim to reduce its complexity~\cite{zhao2023fast,zhang2023faster,xiong2024efficientsam,zhou2023edgesam,shu2025tinysam} or apply SAM to different scenarios such as the medical image field~\cite{mazurowski2023segment,ma2024segment,wu2025medical}.}
SAM's powerful capabilities in image segmentation and its flexible prompting methods have led to its extension to video segmentation.
SAM 2~\cite{ravi2024sam2segmentimages} extends this task to video, known as Promptable Visual Segmentation (PVS). SAM 2 shares the same foundational architecture as SAM, consisting of an image encoder and a prompt-based mask decoder, but SAM 2 achieves video segmentation by incorporating a memory module that conditions the current frame embedding on information from past frames. SAM 2 demonstrates remarkable zero-shot transfer performance and exhibits high versatility across many visual tasks. {However, the complexity in the newly introduced memory modules in SAM 2 needs further reduction. This motivates us to} propose a novel plug-and-play module for memory compression.

\noindent\textbf{Track Anything Model ({TAM}). }
Inspired by the Segment Anything task, the Track Anything paradigm has emerged, which supports flexible selection of target objects and arbitrary adaptation to various video types, enabling multiple downstream video object segmentation (VOS) tasks. 
TAM~\cite{yang2023trackanythingsegmentmeets} combines SAM and XMem~\cite{10.1007/978-3-031-19815-1_37} for interactive video object tracking and segmentation, where SAM is responsible for frame segmentation, and XMem handles tracking. 
The latest SAM 2~\cite{ravi2024sam2segmentimages} extends SAM to video segmentation by introducing a memory module that adjusts the embedding of the current frame based on conditions from past frames. 
\deleted[id=dzy0515]{To enhance the performance, DAM4SAM~\cite{videnovic2025distractor} introduces a distractor-aware memory module that splits the memory bank into a recent-apearance memory and a distractor-resolving memory. SAM2Long~\cite{ding2025sam2long} proposes a training-free long-term VOS method which employs an object-aware memory modulation. }
To meet practical deployment requirements, follow-up work continues to optimize efficiency: EdgeTAM~\cite{Zhou_2025_CVPR} compresses frame-level memories using a 2D spatial transformer for resource-constrained scenarios, while EfficientTAM~\cite{xiong2024efficienttrack} achieves faster cross-attention mechanisms by leveraging the locality of spatial memory embeddings. SAM-I2V~\cite{Mei_2025_CVPR} and others have further advanced this field by reducing training costs.
\section{Method} \label{sec:method}

\subsection{Preliminary: Segment Anything Model 2 (SAM 2)}

The architecture of SAM 2~\cite{ravi2024sam2segmentimages} is optimized for real-time video segmentation and object tracking. SAM 2 inherits the hierarchical image encoder, prompt encoder, and mask decoder components from SAM, and introduces a new memory mechanism that includes a memory attention module, a lightweight memory encoder and a lightweight memory bank.

The memory attention module $A$ consists of a series of {stacked transformer} blocks. Each block includes self-attention, cross-attention, and an MLP, which {interact} the features of the current frame $F_{curr}$ with the memory frame features $M_{i=1}^{t}$ {in the memory bank that} contain the features of previous frames and corresponding predictions:  
\begin{equation}
	F_{attn} = A\left( F_{curr}, M_1,\dots,M_t \right),
	\label{eq-3-1-1}
\end{equation}

Among these, $F_{attn}$ represents the image features generated based on memory conditions. The introduced memory mechanism enables SAM 2 to support real-time video segmentation and object tracking, thus addressing the challenges posed by dynamic scenes. Next, the mask decoder $D_{mask}$ decodes the attention features $F_{attn}$, the current frame image features $F_{curr}$, and the user prompt $P$ to produce the segmentation mask prediction $S$:

\begin{equation}
	S = D_{mask}\left(F_{attn}, F_{curr}, P \right),
	\label{eq-3-1-2}
\end{equation}

The memory encoder $E_{mem}$ encodes the current image features {$F_{curr}$} and the current mask predictions {$S$} together, and {the resulted memory feature $M_{t+1}$ is pushed} into the memory bank in a first-in-first-out manner.
\begin{equation}
	M_{t+1} = E_{mem}\left( F_{curr},S\right),
	\label{eq-3-1-3}
\end{equation}
The memory bank $\mathcal{M}_s \in \mathbb{R}^{t \times h \times w \times c}$ contains the user-provided initial frame and its segmentation mask, and adopts a variant of the memory management mechanism proposed in ~\cite{Zhou_2024_CVPR}. The initial {prompted} frame {(GT frame)} is always retained in the memory bank, while the six most recent frames are updated using a first-in-first-out protocol whenever a new frame arrives.

\begin{figure*}[!ht]
    
    \includegraphics[height=5.4cm ]{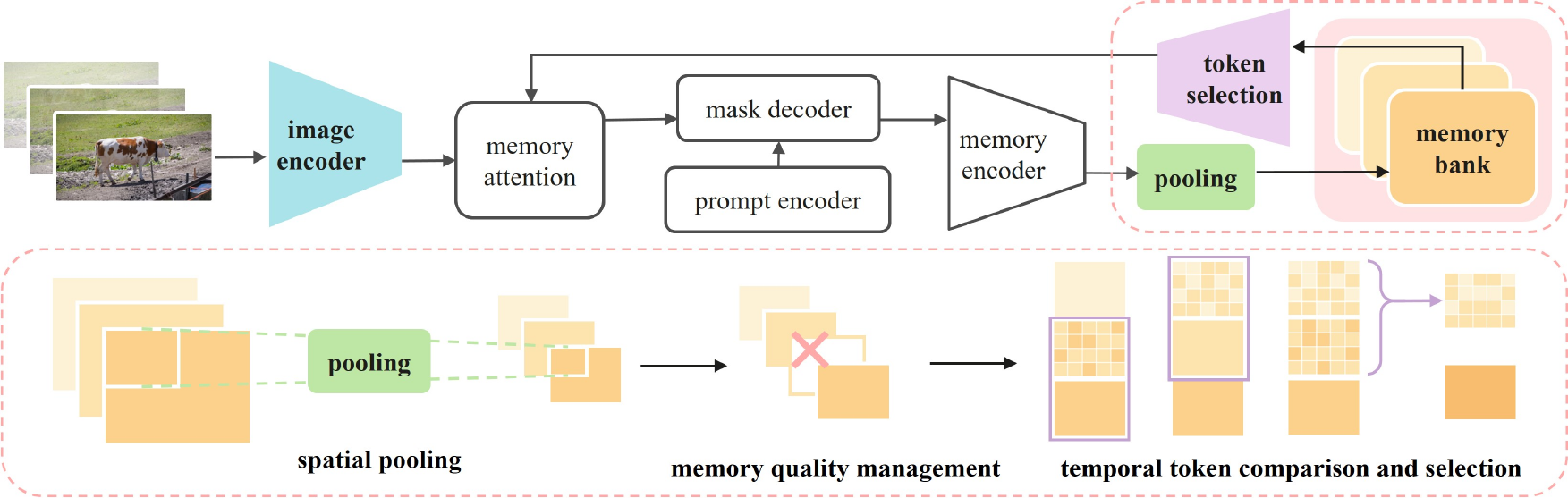}
    \caption{\textbf{Overall architecture of TinySAM 2.} Our work focuses on the colored modules in the diagram. We adopt a lighter image encoder based on SAM 2 to reduce the number of parameters. A spatial pooling compression is performed on individual memory frames before passing to the memory bank, and a temporal token compression module is proposed to compress tokens across multiple frames by the token selection. In addition, we implement quality management for the construction of memory bank.}
	\vspace{-.3cm}
    \label{fig3}
\end{figure*}

{
\subsection{TinySAM 2}
The memory attention module introduced in SAM 2 involves cross-attention between the current frame features and the memory frame features in the memory bank. Thus, the heavy memory features produce a high computational complexity. To address this concern, we propose TinySAM~2, a lightweight solution which introduces memory quality management and memory token compression into the process. The overall framework is illustrated in Figure~\ref{fig3}. Our proposed compression process performs the spatial and temporal compression separately to the memory frames, which is inspired by the video compression paradigm. Specificallly, the frames in the memory bank are treated as two categories: reference frame (GT frame) and motion frames, where spatial memory compression is performed on all the frames, while the temporal memory compression is done only among the motion frames. Besides, a memory management method is proposed to build a qualified memory bank. We use RepViT~\cite{Wang_2024_CVPR} as the lightweight image encoder to further reduce the model parameters.
}

\subsubsection{Spatial Memory Compression}
Considering the local smoothness 
of individual memory frames {in the spatial domain}~\cite{xiong2024efficienttrack}, we employ average pooling to perform coarse-grained compression on single-frame features, allowing adjacent and highly correlated labels to be aggregated. Specifically, for a memory frame label $M = [m_{11},\dots,m_{1h};\dots;m_{w1},\dots,m_{wh}]$ with resolution $w\times h$, we apply pooling over local regions using a sliding window with size $\Delta{w}\times\Delta{h}$ and stride $\Delta{w}\times\Delta{h}$. In this case, the spatial structure is scaled down to $\hat{w}\times\hat{h}=w\times h/\Delta{w}\Delta{h}$, where generally $\Delta{w}=2, \Delta{h}=2$. {Therefore}, the final number of {tokens each} memory frame $M$ obtained is $hw/4$. This operation is independently applied to each memory frame once, and the resulting compressed labels are directly stored in the memory bank. This approach significantly reduces storage and subsequent computational costs while preserving the key spatial structures within {each frame}.

\subsubsection{Temporal Memory Compression}
In a memory bank composed of multiple consecutive frames, there exists significant temporal redundancy. Specifically, background regions and static objects often remain highly consistent across adjacent frames, resulting in extremely subtle visual changes~\cite{Tao_2025_CVPR}. To reduce the computational cost in the cross-attention mechanism, such redundancies can be merged along the temporal dimension.

We draw inspiration from the concepts of frame types in video compression. We treat the ground-truth (GT) frame as an intra-coded frame (I frame), considering it contains the most important appearance information for object tracking. All the tokens are retained after its own spatial compression, serving as the starting point for mask generation. We treat other incoming frames in the memory bank as forward predictive frames (P frames) that concentrate more on the motion-relevant regions. Forward motion compensation prediction is performed based on the previous frame to eliminate temporal redundancy. 
Specifically, we perform continuous sampling on the visual features 
and use a sliding window mechanism with a length of two frames to calculate the similarity between spatially corresponding features 
in adjacent frames. Here, we employ cosine similarity as the metric, defining the similarity score $s$ as:

\begin{equation}
	s=\frac{M_t \cdot M_{t+1}}{||M_t||\cdot||M_{t + 1}||},
	\label{eq-3-4-1}
\end{equation}

{All the tokens in the ground-truth frame $M_{gt}$ are retained as the reference appearance features. Among the remaining memory frames, we then perform global sampling based on all similarity scores.}
Let the number of samples be $n$, then the selected set of tokens and the final set of tokens can be re-expressed as:
\begin{equation}
	M_{sel} = \text{Top-}n (\{m_i\}_{i=1}^{(t-1)\times \hat{h} \times \hat{w} }),
	\label{eq-3-4-2}
\end{equation}

\begin{equation}
	\mathcal{M}_{t} = \left[M_{gt}, M_{sel}\right],
	\label{eq-3-4-3}
\end{equation}

Among them, Top-$n$ indicates selecting the top $n$ tokens in ascending order of similarity, where generally $n=\hat{w}\times \hat{h}$. Through this strategy, we effectively compress temporal redundancy, significantly reducing the computational complexity of the subsequent cross-attention module while preserving key visual information.

\subsubsection{Memory Quality Management}
In memory-based methods, the memory management mechanism is one of the key factors affecting performance. To efficiently utilize limited memory resources, it is essential to prioritize the quality of frames in the memory bank rather than simply pursuing frame continuity. 
Existing methods such as SAM 2~\cite{ravi2024sam2segmentimages} include every frame in the memory bank regardless of its quality. However, this strategy has obvious drawbacks: when the target is briefly occluded, the memory bank can quickly be filled with a large number of frames that do not contain the target, thereby diluting the diversity of the target appearance features and weakening the model's discriminative ability when the target reappears, leading to a decline in segmentation accuracy~\cite{Videnovic_2025_CVPR}. In addition, if the target is not successfully detected, the memory bank may be erroneously updated with empty masks, leading to error accumulation and even permanent loss of the target. 

To alleviate these issues, we propose a quality-based memory bank update strategy: {a mixture of object existence measurement and confidence measurement. First,} when the target is absent in the current frame, the frame is filtered out and will not be included in the memory bank. 
{In that case, we continue} searching for subsequent frames to maintain the dynamic update of the memory bank. In addition, the quality of mask generation also has a significant impact on tracking stability. Low-quality masks usually indicate that the model lacks confidence in its predictions for the frame, which can easily lead to incorrect tracking or ambiguity, and thus result in error accumulation. Therefore, to further ensure the reliability of frames in the memory bank, 
{a frame is included in the memory bank only if} the IoU score predicted 
is above a preset threshold ($\theta$ = 0.5). Otherwise, the frame is skipped, and the subsequent frames are processed.   
Through these strategies, the contamination of low-quality frames in the memory bank can be effectively suppressed.




\section{Experiments} \label{sec:experiments}
\subsection{Experimental Setting}

\noindent\textbf{Training Configurations.}
We set the input resolution to $768\times768$ to make use of limited memory resources.
We employ the AdamW~\cite{loshchilov2018decoupled} optimizer ($\beta_1=0.9$, $\beta_2=0.999$), and apply $L_2$ gradient clipping with a threshold of 0.1 and set weight decay to 0.1. 
The number of training masks per image is limited to 32. The learning rate for the image encoder and other components is set to $6e^{-5}$ with a cosine scheduler. 
Each video sample contains nearly 3 objects, and is augmented with horizontal flipping, color jittering, affine transformations, and grayscale transformations.
We supervise mask predictions using a linear combination of focal loss and Dice loss, while mean absolute error (MAE) loss is used for IoU predictions, and cross-entropy loss for object predictions, with weight ratios of 20:1:1:1, respectively. 
Bfloat16 format was used during training. Our model is pre-trained on 80 GPUs, each with 32GB of VRAM.
We use the RepViT model~\cite{Wang_2024_CVPR} as the image encoder, which has 1/5 of the parameters compared to the original Hiera-Tiny~\cite{ravi2024sam2segmentimages}.

\noindent\textbf{Baseline and Evaluation Metrics.}
Same as the others, the ground truth masks of the first frame are available during inference under semi-supervised VOS scenario. 
We compare the performance of TinySAM 2 with methods such as STCN~\cite{10.5555/3540261.3541162}, RDE~\cite{Li_2022_CVPR}, XMem~\cite{10.1007/978-3-031-19815-1_37},  DEVA~\cite{Cheng_2023_ICCV}, Cutie-base~\cite{Cheng_2024_CVPR}, Cutie-base+~\cite{Cheng_2024_CVPR}, SAM 2.1(Hiera-T)~\cite{ravi2024sam2segmentimages}, EfficientTAM~\cite{xiong2024efficienttrack}, EdgeTAM~\cite{Zhou_2025_CVPR}, SAM-I2V~\cite{Mei_2025_CVPR}. 
In these evaluations, we use $\mathcal{J \& F}$~\cite{ponttuset20182017davischallengevideo} as the evaluation metric, which combines region similarity (Jaccard index) and boundary accuracy (F-measure). FPS is measured on a single GPU with 32GB memory.

\noindent\textbf{Datasets.}
We train on the SA-V~\cite{ravi2024sam2segmentimages}, DAVIS~\cite{ponttuset20182017davischallengevideo}, YTVOS datasets~\cite{xu2018youtubevoslargescalevideoobject} and a 0.5\% subset of SA-1B~\cite{Kirillov_2023_ICCV} for 33 epochs. 
SA-1B contains 11 million images with 1.1 billion mask annotations at different granularities, and the average resolution of images in SA-1B is 3300×4950 pixels. To date, it is the largest image segmentation task dataset available. 
SA-V follows the same standards as SA-1B and collects 190,900 masklets annotations and 600,000 mask annotations from 50,900 videos. These videos range in length from 4 seconds to 138 seconds, with an average duration of 14 seconds. The indoor/outdoor scene ratio is 54\%/46\%, and the videos are resampled to 24 FPS. It is important to note that the annotation frame rate is 6 FPS. 
Accordingly, we provide performance results on the DAVIS 2017~\cite{ponttuset20182017davischallengevideo} and YTVOS ~\cite{xu2018youtubevoslargescalevideoobject} validation sets, as well as the challenging SA-V validation/test sets ~\cite{ravi2024sam2segmentimages}. 
SA-V retains 293 masklets annotations from 155 videos for the validation set and 278 masklets annotations from 150 videos for the test set. These annotations were manually selected to focus on challenging cases involving fast motion, complex occlusions, and disappearances.
\begin{table*}[!ht]
  \caption{Results in the Standard Semi-Supervised Video Object Segmentation Benchmarks.}
  \label{tab:c4-1-compare-to-SOTA}
  \centering
  \begin{tabular}{lccccccccc}
    \toprule
    \multirow{3}{*}{Method} 
    & \multicolumn{5}{c}{$\mathcal{J\& F} \uparrow$ } 
    & \multirow{3}{*}{\begin{tabular}[c]{@{}c@{}}Parameters\\(M) $\downarrow$\\ \end{tabular}}
    & \multirow{3}{*}{\begin{tabular}[c]{@{}c@{}}Training data \\ ratio $\downarrow$\\ \end{tabular}} 
    & \multirow{3}{*}{\begin{tabular}[c]{@{}c@{}}Tokens \\ ratio $\downarrow$\\ \end{tabular}} 
    & \multirow{3}{*}{\begin{tabular}[c]{@{}c@{}}FPS \\ V100 $\uparrow$\\ \end{tabular}}
    \\
    \cmidrule(lr){2-6}
    & \multirow{2}{*}{\begin{tabular}[c]{@{}c@{}}DAVIS\\2017 val\\ \end{tabular}}
    & \multirow{2}{*}{\begin{tabular}[c]{@{}c@{}}YTVOS\\2019 val\\ \end{tabular}}
    & \multirow{2}{*}{\begin{tabular}[c]{@{}c@{}}SA-V\\ val\\ \end{tabular}}
    & \multirow{2}{*}{\begin{tabular}[c]{@{}c@{}}SA-V\\test\\ \end{tabular}}
    & \multirow{2}{*}{\begin{tabular}[c]{@{}c@{}}avg \end{tabular}} \\ \\
    \midrule
    STCN~\cite{10.5555/3540261.3541162} & 85.4 & 82.7 & 61.0 & 62.5 & 72.9 & 54.0  & 0.02 & 1 & 13.2 \\
    RDE~\cite{Li_2022_CVPR}  & 84.2 & 81.9 & 51.8 & 53.9 & 68.0 & 64.0 & 0.02 & 1 &  - \\
    XMem~\cite{10.1007/978-3-031-19815-1_37} & 86.0 & 85.6 & 60.1 & 62.3 & 73.5 & 62.0 & 0.02 & 1 & - \\
    DEVA~\cite{Cheng_2023_ICCV} & 87.0 & 82.6 & 55.4 & 56.2 & 70.3 & 69.0 & 0.02 & 1 & - \\
    Cutie-base~\cite{Cheng_2024_CVPR}  & 87.9 & 87.0 & 60.7 & 62.7 & 74.6 & 35.0 & 0.02 & 1 & - \\
    Cutie-base+~\cite{Cheng_2024_CVPR} & 88.1 & 87.5 & 61.3 & 62.8 & 75.0 & 35.0 & 0.02 & 1 & 17.9 \\
    SAM 2.1 (T)~\cite{ravi2024sam2segmentimages} & 89.4 & 87.4 & 75.2 & 76.5 & 82.1 & 38.9 & 1 & 1 & 8.8 \\
    EfficientTAM~\cite{xiong2024efficienttrack} & 88.4 & 87.1 & 71.3 & 70.8 & 79.4 & 18.0 & 0.92 & 0.25 & 20.1 \\
    EdgeTAM~\cite{Zhou_2025_CVPR} & 87.7 & 86.2 & 72.3 & 71.7 & 79.5 & 13.9 & 0.92 & 0.14 & 19.2 \\
    SAM-I2V~\cite{Mei_2025_CVPR} & 81.9 & 79.8 & 55.4 & 59.3 & 69.1 & 18.9 & 0.02 & 1 & 6.6 \\
    \textbf{TinySAM 2 (Ours)}  & 83.1 & 81.5 &  65.1 & 65.3 & 73.8 & 17.5 & 0.03 & \textbf{0.07} & \textbf{25.6}\\
    \bottomrule
  \end{tabular}
\end{table*}

\begin{figure*}[!ht]
    
    \rotatebox{90}{\includegraphics[width=6cm]{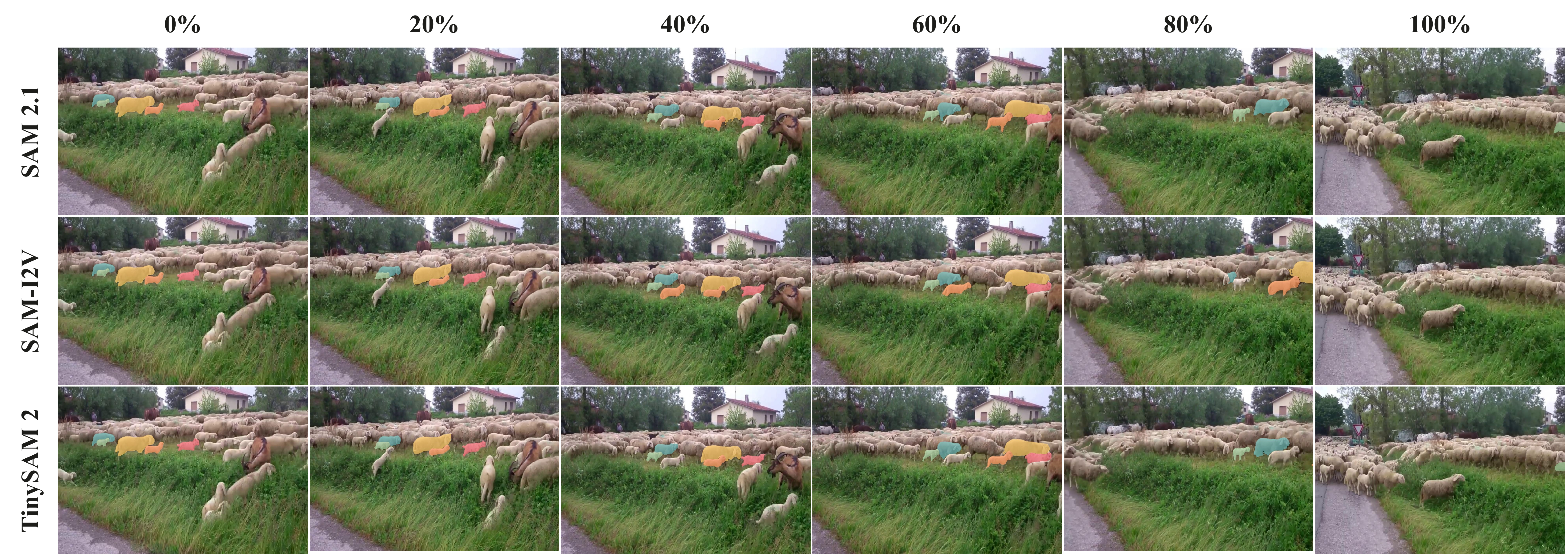}}
    \caption{\textbf{Qualitative comparison of TinySAM 2, SAM 2.1 and SAM-I2V in multi-instance segmentation.} We visually demonstrate the results by uniformly sampling frames from a complete video that tracks five sheep of different sizes with alternating positions within a flock. Our TinySAM 2 generates masks of comparable quality to SAM 2.1, whereas SAM-I2V encounters tracking errors.}
    \vspace{-.3cm}
    \label{fig_c4_quanlitavie_results}
\end{figure*}

\begin{figure*}[!ht]
    
    \centering
    \rotatebox{90}{\includegraphics[width=5.5cm]{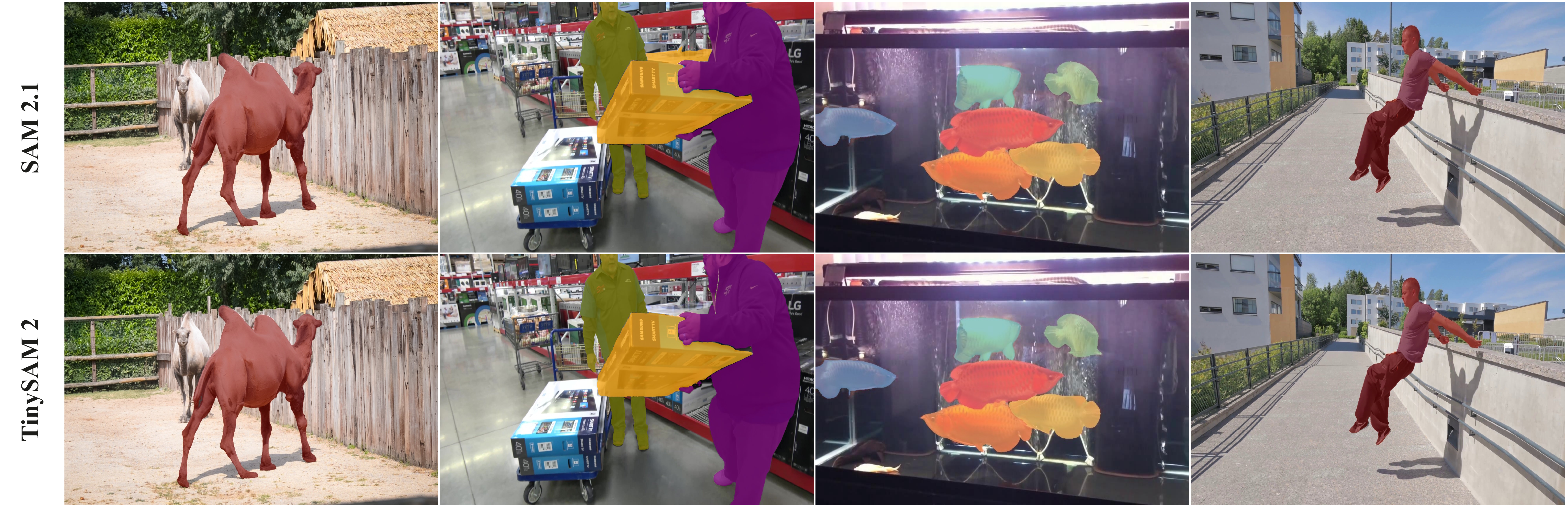}}
    \caption{\textbf{Qualitative comparison of TinySAM 2 and SAM 2.1 across different instances and scenarios.} We visualize frames from different videos involving single-person/multi-person scenes, single-animal/multi-animal scenes, and indoor/outdoor scenes, where TinySAM~2 performs comparably to SAM 2.1.}
    \label{fig_c4_quanlitavie_results2}
\end{figure*}

\subsection{Comparisons with State-of-the-arts}

\textbf{Quantitative Results.}
To assess the effectiveness of our proposed TinySAM 2, we compared it to 10 state-of-the-art methods on common datasets, and the results are reported in Table~\ref{tab:c4-1-compare-to-SOTA}.
The memory reading operation of the STCN needs to calculate the similarity between the memory key and the query key. The cosine similarity or negative squared Euclidean distance similarity involves matrix multiplication. Thus, when the number of tokens in the memory key reaches $thw$, the computational cost is relatively high. The subsequent work RDE, XMem, DEVA, Cuite-base, Cuite-base+ also follows the memory reading operation using negative squared Euclidean distance similarity, and the number of tokens involved is as high as $thw$. In contrast, our TinySAM 2 scores better than STCN, RDE, XMem, and DEVA with fewer parameters (less than 17.5 / 35 = 50\%) and fewer token lengths (7\%), achieved 99\% / 98\% of Cutie-base/Cutie-base++.\looseness=-1

Although SAM 2.1 is a strong baseline model with an average accuracy of 82.1, it requires a large number of computing resources and has a large number of parameters and token calculations. In contrast, our TinySAM 2 achieves an average score of 90\% with only 3\% training data at 3.8\% training cost (i.e., 76.8K G·Hour vs. 2.2M G·Hour). The number of model parameters is reduced by 55\% (i.e. 17.5M vs. 38.9M), and the number of memory tokens is reduced by 93\% (i.e. $hw/2$ vs. $thw$, {where $t=7$}).
As for the improved methods of SAM 2, EfficientTAM and EdgeTAM, the cross-attention operations also involve matrix multiplication between memory key and query key, but the number of tokens in the memory key is reduced to $1/4$ and $1/7$ of the original number, respectively. But our TinySAM 2 achieved an average score of 92.9\% of their performance, with tokens of only 29\% / 57\%.
Compared with SAM-I2V, TinySAM 2 outperforms SAM-I2V in all datasets, especially in SA-V by 9 points, where the memory token length we use is only 7\%. Besides, the inference speed of TinySAM 2 is much faster, that can run in real-time at 25.6 FPS on a single GPU 
\added[id=dzy]{for 1000 image frames from three videos,}
which significantly surpasses SAM-I2V by 3.9 times.
The above results confirm that our TinySAM 2 has a performance close to or even significantly better than some of the existing methods at a much lower cost.

\noindent\textbf{Qualitative Results.} The qualitative comparison results are shown in Figure~\ref{fig_c4_quanlitavie_results}, which presents the visual segmentation results of SAM 2.1, SAM-I2V, and TinySAM 2 in {multi-instance} and occlusion scenarios. All displayed video frames are uniformly extracted with an equal interval from the full segmentation results to ensure the scene variety and objectivity.

From the overall visual results, the three models show relatively small differences in segmentation performance during the early stages of the video, where all models are capable for basic mask prediction of the target objects. However, as the video sequence progresses, performance differences among the models gradually become apparent. 
{On one hand, compared to SAM-I2V, TinySAM 2 demonstrates a significant advantage in instance discrimination and tracking accuracy. Particularly, during object motion and occlusion scenarios, TinySAM 2 can consistently maintain the continuity of tracking target instances, effectively avoiding the instance confusion or tracking loss issues. For instance, starting from the 40\% column in Figure~\ref{fig_c4_quanlitavie_results}, SAM-I2V mistracked the orange sheep in the middle of the scene, which leads to further error accumulation in the rest of the video. Our proposed TinySAM 2 can accurately track each instance, which benefits from memory quality measurement and memory token selection. On the other hand, our proposed TinySAM 2 slightly underperforms SAM 2.1 in fine details like the sheep legs.}
In Figure~\ref{fig_c4_quanlitavie_results2}, we have additionally presented comparisons between TinySAM 2 and SAM 2.1 across more different scenarios, showing that the overall performance is comparable to that of SAM 2.1.

In summary, based on the qualitative analysis results, it is evident that when dealing with complex scenarios such as multi-instance and occlusion, TinySAM 2 not only outperforms SAM-I2V in segmentation and tracking capabilities, but also achieves performance comparable to SAM 2.1. This provides a solution for complex video segmentation tasks that balances both performance and practicality.

\subsection{Ablations}

\begin{table}
  \caption{Ablation on the effectiveness of spatial memory compression.}
  \label{tab:c4-2-1-spatial}
  \centering
  \begin{tabular}{l@{\hspace{4pt}}c@{\hspace{4pt}}c@{\hspace{4pt}}c@{\hspace{4pt}}c@{\hspace{4pt}}c}
    \toprule
    \multirow{3}{*}{Method}

    & \multicolumn{5}{c}{$\mathcal{J\& F} \uparrow$ } \\

    \cmidrule(lr){2-6}

    & \multirow{2}{*}{\begin{tabular}[c]{@{}c@{}}DAVIS\\2017 val\\ \end{tabular}}
    & \multirow{2}{*}{\begin{tabular}[c]{@{}c@{}}YTVOS\\2019 val\\ \end{tabular}}
    & \multirow{2}{*}{\begin{tabular}[c]{@{}c@{}}SA-V\\ val\\ \end{tabular}}
    & \multirow{2}{*}{\begin{tabular}[c]{@{}c@{}}SA-V\\test\\ \end{tabular}}
    & \multirow{2}{*}{\begin{tabular}[c]{@{}c@{}}avg \end{tabular}} \\ \\
    
    \midrule
    SAM 2.1 w/ SMC & 85.5  & 86.0  & 66.8  & 69.1  & 76.9  \\
    \multirow{2}{*}{\makecell{SAM 2.1 w/ SMC \\ \& RepViT}}
    & \multirow{2}{*}{81.5}
    & \multirow{2}{*}{81.3}
    & \multirow{2}{*}{62.6}
    & \multirow{2}{*}{63.2}
    & \multirow{2}{*}{72.2} \\
    & & & & & \\

    \midrule
    Convolution & 81.8  & 80.8  & 63.5  & 65.0  & 72.8   \\
    Max Pooling & 81.3  & \textbf{81.5}  & 65.0  & \textbf{65.8}  & 73.4   \\ 
    \textbf{Average Pooling}  & \textbf{83.1}  & \textbf{81.5}  &  \textbf{65.1}  & 65.3  & \textbf{73.8}  \\
    \bottomrule
  \end{tabular}
\end{table}

\begin{table}
  \caption{Ablation on the effectiveness of memory quality management.}
  \label{tab:c4-2-2-memory-quality}
  \centering
  \vspace{-.2cm}
  \begin{tabular}{l@{\hspace{4pt}}c@{\hspace{4pt}}c@{\hspace{4pt}}c@{\hspace{4pt}}c@{\hspace{4pt}}c@{\hspace{4pt}}c@{\hspace{4pt}}c@{\hspace{4pt}}c@{\hspace{4pt}}c@{\hspace{4pt}}c@{\hspace{4pt}}c@{\hspace{4pt}}c@{\hspace{4pt}}c@{\hspace{4pt}}c@{\hspace{4pt}}c}
    \toprule
    \multirow{3}{*}{Method}

    & \multicolumn{5}{c}{$\mathcal{J\& F} \uparrow$ } \\

    \cmidrule(lr){2-6}

    & \multirow{2}{*}{\begin{tabular}[c]{@{}c@{}}DAVIS\\2017 val\\ \end{tabular}}
    & \multirow{2}{*}{\begin{tabular}[c]{@{}c@{}}YTVOS\\2019 val\\ \end{tabular}}
    & \multirow{2}{*}{\begin{tabular}[c]{@{}c@{}}SA-V\\ val\\ \end{tabular}}
    & \multirow{2}{*}{\begin{tabular}[c]{@{}c@{}}SA-V\\test\\ \end{tabular}}
    & \multirow{2}{*}{\begin{tabular}[c]{@{}c@{}}avg \end{tabular}} \\ \\
    \midrule
    w/o IOU & 81.5  & 81.4  & 64.7  & 64.7  &  73.1   \\ 
    w/o abs. & 82.3  & 81.6  & \textbf{65.3}  & 64.0  & 73.3 \\
    w/o abs. or IOU & 82.9  & \textbf{81.7}  & 64.0   & 64.7  & 73.3 \\
    \textbf{w/ abs. \& IOU} & \textbf{83.1}   & 81.5  &  65.1  & \textbf{65.3}  & \textbf{73.8}  \\
    \bottomrule
  \end{tabular}
\end{table}

\begin{table}[t]
  \caption{\textbf{Ablation on the effectiveness of temporal memory compression.} \enquote{GT/first + last} indicates retaining the GT or the first and the last frame in the memory bank, totaling 2 frames.\added[id=dzy]{\enquote{w/o TMC} indicates that the TMC module is removed and the memory frames that are not compressed in the time sequence are used.} \enquote{GT/prev. + frame/global} means comparing each frame with the GT/previous frame to measure the similarity, and then uniformly selecting tokens from each frame or selection tokens globally across frames.}
  \label{tab:c4-2-3-temporal}
  \centering
  \vspace{-.2cm}

  \begin{tabular}{l@{\hspace{5pt}}c@{\hspace{5pt}}c@{\hspace{5pt}}c@{\hspace{5pt}}c@{\hspace{5pt}}c@{\hspace{5pt}}c@{\hspace{5pt}}c@{\hspace{5pt}}c@{\hspace{5pt}}c@{\hspace{5pt}}c@{\hspace{5pt}}c@{\hspace{5pt}}c@{\hspace{5pt}}c@{\hspace{5pt}}c@{\hspace{5pt}}c}
    \toprule
    \multirow{3}{*}{Method}

    & \multicolumn{5}{c}{$\mathcal{J\& F} \uparrow$ } \\

    \cmidrule(lr){2-6}

    & \multirow{2}{*}{\begin{tabular}[c]{@{}c@{}}DAVIS\\2017 val\\ \end{tabular}}
    & \multirow{2}{*}{\begin{tabular}[c]{@{}c@{}}YTVOS\\2019 val\\ \end{tabular}}
    & \multirow{2}{*}{\begin{tabular}[c]{@{}c@{}}SA-V\\ val\\ \end{tabular}}
    & \multirow{2}{*}{\begin{tabular}[c]{@{}c@{}}SA-V\\test\\ \end{tabular}}
    & \multirow{2}{*}{\begin{tabular}[c]{@{}c@{}}avg \end{tabular}} \\ \\
    \midrule

    First + Last & 80.6  & 79.7  & 55.0  & 56.9  & 68.1  \\
    GT + Last  & 82.3  & 80.4  & 61.9  & 63.8  & 72.1  \\
    w/o TMC & 83.0 & \textbf{81.7} & 63.1 & 65.3 & 73.3 \\
    GT + Frame & 81.5 & \textbf{81.7}  & 65.3  & 66.0 & 73.6  \\
    Prev. + Frame & 81.6  & 81.5 & 65.0 & 65.8 & 73.5 \\
    GT + Global & 81.5 & 81.2 & \textbf{66.3} & \textbf{66.1} & \textbf{73.8} \\
    \textbf{Prev.} + \textbf{Global} & \textbf{83.1}  & 81.5 &  65.1  & 65.3  & \textbf{73.8}  \\
    \bottomrule
  \end{tabular}
\end{table}

\begin{table}
  \caption{\textbf{Ablation on the effectiveness of different compression ratio.} \enquote{28:3} indicates retaining 3 frames (GT, first, last), and \enquote{28:1} indicates reducing $t=7$ frames to 1 frame through a moving average. The compression ratio takes the spatial pooling by a factor of 4 into account as well.}
  \label{tab:c4-2-4-ratio}
  \centering
	\vspace{-.25cm}
  \begin{tabular}{lc@{\hspace{7pt}}c@{\hspace{7pt}}c@{\hspace{7pt}}c@{\hspace{7pt}}c@{\hspace{7pt}}c@{\hspace{7pt}}c@{\hspace{7pt}}c@{\hspace{7pt}}c@{\hspace{7pt}}c@{\hspace{7pt}}c@{\hspace{7pt}}c@{\hspace{7pt}}c@{\hspace{7pt}}c@{\hspace{7pt}}c}
    \toprule
    \multirow{3}{*}{Method}

    & \multicolumn{5}{c}{$\mathcal{J\& F} \uparrow$ } \\

    \cmidrule(lr){2-6}

    & \multirow{2}{*}{\begin{tabular}[c]{@{}c@{}}DAVIS\\2017 val\\ \end{tabular}}
    & \multirow{2}{*}{\begin{tabular}[c]{@{}c@{}}YTVOS\\2019 val\\ \end{tabular}}
    & \multirow{2}{*}{\begin{tabular}[c]{@{}c@{}}SA-V\\ val\\ \end{tabular}}
    & \multirow{2}{*}{\begin{tabular}[c]{@{}c@{}}SA-V\\test\\ \end{tabular}}
    & \multirow{2}{*}{\begin{tabular}[c]{@{}c@{}}avg \end{tabular}} \\ \\
    \midrule
    
    28:1 & 81.6 & 77.5 & 48.8 & 52.4 & 65.0 \\
    \textbf{14:1}  & \textbf{83.1}  & \textbf{81.5} &  65.1 & 65.3 & 73.8  \\
    
    28:3 & 82.6 & \textbf{81.5} & \textbf{66.4} & \textbf{66.5}  & \textbf{74.3}  \\
    \bottomrule
  \end{tabular}
\end{table}

We conduct extensive ablation experiments using the same training strategy to validate the effectiveness of each key component of TinySAM 2. Tables~\ref{tab:c4-2-1-spatial}, \ref{tab:c4-2-2-memory-quality}, \ref{tab:c4-2-3-temporal}, \ref{tab:c4-2-4-ratio} summarize our findings.

\noindent\textbf{Effectiveness of Spatial Memory Compression.}
\replaced[id=dzy]{In Table~\ref{tab:c4-2-1-spatial}, we compare the impact of spatial memory compression (SMC) and also evaluate several other spatial compression pooling methods. Since SAM 2.1 cannot be simply trained after replacing ViT on a 32GB V100 GPU, we first observe the effect by adding a spatial compression module to SAM 2.1 and then replace ViT based on this observation to evaluate the impact. Compared to SAM 2.1, these two modules significantly reduce the average $\mathcal{J\& F}$ score.}{Several spatial compression pooling methods are compared in Table~\ref{tab:c4-2-1-spatial}.}
\added[id=dzy]{The inference speeds of SAM 2.1 w/ SMC, w/ SMC \& RepViT are 18.2 FPS and 24.9 FPS, respectively, which are significantly higher than the inference speed of SAM 2.1, indicating that the two corresponding modules have a significant acceleration effect on the inference speed.}
\added[id=dzy]{In the exploration of the effectiveness of spatial compression methods in TinySAM 2}
 the average pooling TinySAM 2 used achieves the best $\mathcal{J\& F}$ score in most cases as well as the averaging. Max pooling achieves slightly lower scores.
Notably, although convolution introduces more parameters, its performance on all datasets is inferior to that of average pooling.

\noindent\textbf{Effectiveness of Memory Quality Management.}
In Table~\ref{tab:c4-2-2-memory-quality}, we investigate the impact of memory quality management. Our method improves the score by 0.7 points compared to the lowest-scoring w/o IOU, and different methods show varying performance across datasets, indicating that the absence of tracked objects varies among datasets. However, on average, memory quality management generally led to a certain improvement. \added[id=dzy]{In terms of inference speed, the speed of w/o absence or IOU is 26.2 FPS, which is higher than the inference speed of TinySAM 2. This indicates that this module may cause a decrease in inference speed.}

\noindent\textbf{Effectiveness of Temporal Memory Compression.}
We compare several methods for compressing tokens along the temporal dimension. In Table~\ref{tab:c4-2-3-temporal}, we find that simply retaining only the first and last frames in the memory bank yielded the worst performance, with an averaged $\mathcal{J\& F}$ score of only 68.1. 
Retaining the GT frame and the nearest frame significantly improves the score to 72.1. 
Further improvement is achieved by replacing the nearest frame with our proposed method of selecting tokens based on cosine similarity after contrastive comparison. Notably, calculating similarity between each frame and its previous frame or between each frame and the GT frame yielded comparable results, while selecting tokens globally between frames is slightly better than the method of uniformly selecting tokens from each frame.
\added[id=dzy]{Our temporal memory compression method also outperforms the uncompressed sequence w/o TMC, and the inference speed of w/o TMC is 17.2 FPS, which is significantly lower than that of TinySAM 2. This indicates that the temporal memory compression method not only improves the performance but also significantly accelerates the inference speed.}

\noindent\textbf{Different Compression Ratio.}
In Table~\ref{tab:c4-2-4-ratio}, we explore the effects of different compression ratios.
 A higher compression ratio 28:1 and a lower compression ratio 28:3 are examined. It shows that further increasing the compression ratio leads to a noticeable decline in scores, with the $\mathcal{J\& F}$ average score dropping by more than 10 percentage points. Conversely, reducing the compression ratio results in a slight improvement in scores, with the $\mathcal{J\& F}$ average score increased by 0.5 percentage points. This indicates that the current method has achieved a good balance between compression rate and performance.

\section{Conclusion} \label{sec:conclusion}
In this paper, we proposed TinySAM 2, which is a lightweight and efficient track anything model that employs memory quality management and memory token compression. The memory quality management enables the selection and retention of high-quality memories, laying a foundation for efficient operation. To further reduce computational overhead, {spatiotemporal compression was introduced to significantly decrease the amount of memory tokens processed in the memory attention. Two techniques were proposed in the compression scheme: the intra-frame spatial pooling and the inter-frame temporal token comparison based on similarity measurement.} 
Comprehensive experimental results validate the superiority of TinySAM 2, which requires only {$7\%$ of memory tokens that achieves at an extreme memory compression and can run in real-time. At the same time, TinySAM 2 needs only $3\%$ of the training data compared with SAM 2.1}, greatly reducing computations and improving inference efficiency. 
Despite the substantial reduction in both parameters and computational cost, TinySAM 2 achieves 90\% of the performance of SAM 2.1, demonstrating an excellent balance between performance and efficiency.
{
    \small
    \bibliographystyle{ieeenat_fullname}
    \bibliography{tinysam2_sec/tinysam2}

@String(CVPR= {IEEE Conf. Comput. Vis. Pattern Recog.})

@String(ICCV= {Int. Conf. Comput. Vis.})

@String(ECCV= {Eur. Conf. Comput. Vis.})

@String(NIPS= {Adv. Neural Inform. Process. Syst.})

@String(AAAI = {AAAI})

@String(CVPR  = {CVPR})

@String(ICCV  = {ICCV})

@String(ECCV  = {ECCV})

@String(NIPS  = {NeurIPS})

@InProceedings{Zhou_2025_CVPR,
    author    = {Zhou, Chong and Zhu, Chenchen and Xiong, Yunyang and Suri, Saksham and Xiao, Fanyi and Wu, Lemeng and Krishnamoorthi, Raghuraman and Dai, Bo and Loy, Chen Change and Chandra, Vikas and Soran, Bilge},
    title     = {EdgeTAM: On-Device Track Anything Model},
    booktitle = {Proceedings of the IEEE/CVF Conference on Computer Vision and Pattern Recognition (CVPR)},
    month     = {June},
    year      = {2025},
    pages     = {13832-13842}
}

@misc{xiong2024efficienttrack,
      title={Efficient Track Anything}, 
      author={Yunyang Xiong and Chong Zhou and Xiaoyu Xiang and Lemeng Wu and Chenchen Zhu and Zechun Liu and Saksham Suri and Balakrishnan Varadarajan and Ramya Akula and Forrest Iandola and Raghuraman Krishnamoorthi and Bilge Soran and Vikas Chandra},
      year={2024},
      eprint={2411.18933},
      archivePrefix={arXiv},
      primaryClass={cs.CV},
      url={https://arxiv.org/abs/2411.18933}, 
}

@InProceedings{Mei_2025_CVPR,
    author    = {Mei, Haiyang and Zhang, Pengyu and Shou, Mike Zheng},
    title     = {SAM-I2V: Upgrading SAM to Support Promptable Video Segmentation with Less than 0.2\% Training Cost},
    booktitle = {Proceedings of the IEEE/CVF Conference on Computer Vision and Pattern Recognition (CVPR)},
    month     = {June},
    year      = {2025},
    pages     = {3417-3426}
}

@misc{ravi2024sam2segmentimages,
      title={SAM 2: Segment Anything in Images and Videos}, 
      author={Nikhila Ravi and Valentin Gabeur and Yuan-Ting Hu and Ronghang Hu and Chaitanya Ryali and Tengyu Ma and Haitham Khedr and Roman Rädle and Chloe Rolland and Laura Gustafson and Eric Mintun and Junting Pan and Kalyan Vasudev Alwala and Nicolas Carion and Chao-Yuan Wu and Ross Girshick and Piotr Dollár and Christoph Feichtenhofer},
      year={2024},
      eprint={2408.00714},
      archivePrefix={arXiv},
      primaryClass={cs.CV},
      url={https://arxiv.org/abs/2408.00714}, 
}

@InProceedings{Kirillov_2023_ICCV,
    author    = {Kirillov, Alexander and Mintun, Eric and Ravi, Nikhila and Mao, Hanzi and Rolland, Chloe and Gustafson, Laura and Xiao, Tete and Whitehead, Spencer and Berg, Alexander C. and Lo, Wan-Yen and Dollar, Piotr and Girshick, Ross},
    title     = {Segment Anything},
    booktitle = {Proceedings of the IEEE/CVF International Conference on Computer Vision (ICCV)},
    month     = {October},
    year      = {2023},
    pages     = {4015-4026}
}

@misc{ponttuset20182017davischallengevideo,
      title={The 2017 DAVIS Challenge on Video Object Segmentation}, 
      author={Jordi Pont-Tuset and Federico Perazzi and Sergi Caelles and Pablo Arbeláez and Alex Sorkine-Hornung and Luc Van Gool},
      year={2018},
      eprint={1704.00675},
      archivePrefix={arXiv},
      primaryClass={cs.CV},
      url={https://arxiv.org/abs/1704.00675}, 
}

@misc{xu2018youtubevoslargescalevideoobject,
      title={YouTube-VOS: A Large-Scale Video Object Segmentation Benchmark}, 
      author={Ning Xu and Linjie Yang and Yuchen Fan and Dingcheng Yue and Yuchen Liang and Jianchao Yang and Thomas Huang},
      year={2018},
      eprint={1809.03327},
      archivePrefix={arXiv},
      primaryClass={cs.CV},
      url={https://arxiv.org/abs/1809.03327}, 
}

@inproceedings{10.5555/3540261.3541162,
author = {Cheng, Ho Kei and Tai, Yu-Wing and Tang, Chi-Keung},
title = {Rethinking space-time networks with improved memory coverage for efficient video object segmentation},
year = {2021},
isbn = {9781713845393},
publisher = {Curran Associates Inc.},
address = {Red Hook, NY, USA},
booktitle = {Proceedings of the 35th International Conference on Neural Information Processing Systems},
articleno = {901},
numpages = {14},
series = {NIPS '21}
}

@InProceedings{Li_2022_CVPR,
    author    = {Li, Mingxing and Hu, Li and Xiong, Zhiwei and Zhang, Bang and Pan, Pan and Liu, Dong},
    title     = {Recurrent Dynamic Embedding for Video Object Segmentation},
    booktitle = {Proceedings of the IEEE/CVF Conference on Computer Vision and Pattern Recognition (CVPR)},
    month     = {June},
    year      = {2022},
    pages     = {1332-1341}
}

@InProceedings{10.1007/978-3-031-19815-1_37,
    author="Cheng, Ho Kei
    and Schwing, Alexander G.",
    editor="Avidan, Shai
    and Brostow, Gabriel
    and Ciss{\'e}, Moustapha
    and Farinella, Giovanni Maria
    and Hassner, Tal",
    title="XMem: Long-Term Video Object Segmentation with an Atkinson-Shiffrin Memory Model",
    booktitle="Computer Vision -- ECCV 2022",
    year="2022",
    publisher="Springer Nature Switzerland",
    address="Cham",
    pages="640--658",
    isbn="978-3-031-19815-1"
}

@InProceedings{Cheng_2023_ICCV,
    author    = {Cheng, Ho Kei and Oh, Seoung Wug and Price, Brian and Schwing, Alexander and Lee, Joon-Young},
    title     = {Tracking Anything with Decoupled Video Segmentation},
    booktitle = {Proceedings of the IEEE/CVF International Conference on Computer Vision (ICCV)},
    month     = {October},
    year      = {2023},
    pages     = {1316-1326}
}

@InProceedings{Cheng_2024_CVPR,
    author    = {Cheng, Ho Kei and Oh, Seoung Wug and Price, Brian and Lee, Joon-Young and Schwing, Alexander},
    title     = {Putting the Object Back into Video Object Segmentation},
    booktitle = {Proceedings of the IEEE/CVF Conference on Computer Vision and Pattern Recognition (CVPR)},
    month     = {June},
    year      = {2024},
    pages     = {3151-3161}
}

@inproceedings{
    loshchilov2018decoupled,
    title={Decoupled Weight Decay Regularization},
    author={Ilya Loshchilov and Frank Hutter},
    booktitle={International Conference on Learning Representations},
    year={2019},
    url={https://openreview.net/forum?id=Bkg6RiCqY7},
}

@article{Cheng2023PuttingTO,
  title={Putting the Object Back into Video Object Segmentation},
  author={Ho Kei Cheng and Seoung Wug Oh and Brian L. Price and Joon-Young Lee and Alexander G. Schwing},
  journal={2024 IEEE/CVF Conference on Computer Vision and Pattern Recognition (CVPR)},
  year={2023},
  pages={3151-3161},
  url={https://api.semanticscholar.org/CorpusID:264305820}
}

@INPROCEEDINGS{10943417,
  author={Shaker, Abdelrahman and Talal, Syed and Danelljan, Martin and Khan, Salman and Yang, Ming-Hsuan and Khan, Fahad Shahbaz},
  booktitle={2025 IEEE/CVF Winter Conference on Applications of Computer Vision (WACV)}, 
  title={Efficient Video Object Segmentation via Modulated Cross-Attention Memory}, 
  year={2025},
  volume={},
  number={},
  pages={8681-8690},
  doi={10.1109/WACV61041.2025.00841}}

@ARTICLE{10487964,
  author={Yang, Zongxin and Miao, Jiaxu and Wei, Yunchao and Wang, Wenguan and Wang, Xiaohan and Yang, Yi},
  journal={IEEE Transactions on Pattern Analysis and Machine Intelligence}, 
  title={Scalable Video Object Segmentation With Identification Mechanism}, 
  year={2024},
  volume={46},
  number={9},
  pages={6247-6262},
  doi={10.1109/TPAMI.2024.3383592}}

@InProceedings{Wang_2024_CVPR,
    author    = {Wang, Ao and Chen, Hui and Lin, Zijia and Han, Jungong and Ding, Guiguang},
    title     = {RepViT: Revisiting Mobile CNN From ViT Perspective},
    booktitle = {Proceedings of the IEEE/CVF Conference on Computer Vision and Pattern Recognition (CVPR)},
    month     = {June},
    year      = {2024},
    pages     = {15909-15920}
}

@InProceedings{Zhou_2024_CVPR,
    author    = {Zhou, Junbao and Pang, Ziqi and Wang, Yu-Xiong},
    title     = {RMem: Restricted Memory Banks Improve Video Object Segmentation},
    booktitle = {Proceedings of the IEEE/CVF Conference on Computer Vision and Pattern Recognition (CVPR)},
    month     = {June},
    year      = {2024},
    pages     = {18602-18611}
}

@InProceedings{Videnovic_2025_CVPR,
    author    = {Videnovic, Jovana and Lukezic, Alan and Kristan, Matej},
    title     = {A Distractor-Aware Memory for Visual Object Tracking with SAM2},
    booktitle = {Proceedings of the IEEE/CVF Conference on Computer Vision and Pattern Recognition (CVPR)},
    month     = {June},
    year      = {2025},
    pages     = {24255-24264}
}

@InProceedings{Tao_2025_CVPR,
    author    = {Tao, Keda and Qin, Can and You, Haoxuan and Sui, Yang and Wang, Huan},
    title     = {DyCoke: Dynamic Compression of Tokens for Fast Video Large Language Models},
    booktitle = {Proceedings of the IEEE/CVF Conference on Computer Vision and Pattern Recognition (CVPR)},
    month     = {June},
    year      = {2025},
    pages     = {18992-19001}
}

@InProceedings{pmlr-v202-ryali23a,
  title = 	 {Hiera: A Hierarchical Vision Transformer without the Bells-and-Whistles},
  author =       {Ryali, Chaitanya and Hu, Yuan-Ting and Bolya, Daniel and Wei, Chen and Fan, Haoqi and Huang, Po-Yao and Aggarwal, Vaibhav and Chowdhury, Arkabandhu and Poursaeed, Omid and Hoffman, Judy and Malik, Jitendra and Li, Yanghao and Feichtenhofer, Christoph},
  booktitle = 	 {Proceedings of the 40th International Conference on Machine Learning},
  pages = 	 {29441--29454},
  year = 	 {2023},
  editor = 	 {Krause, Andreas and Brunskill, Emma and Cho, Kyunghyun and Engelhardt, Barbara and Sabato, Sivan and Scarlett, Jonathan},
  volume = 	 {202},
  series = 	 {Proceedings of Machine Learning Research},
  month = 	 {23--29 Jul},
  publisher =    {PMLR},
  url = 	 {https://proceedings.mlr.press/v202/ryali23a.html}
}

@misc{yang2023trackanythingsegmentmeets,
      title={Track Anything: Segment Anything Meets Videos}, 
      author={Jinyu Yang and Mingqi Gao and Zhe Li and Shang Gao and Fangjing Wang and Feng Zheng},
      year={2023},
      eprint={2304.11968},
      archivePrefix={arXiv},
      primaryClass={cs.CV},
      url={https://arxiv.org/abs/2304.11968}, 
}

@inproceedings{caelles2017one,
  title={One-shot video object segmentation},
  author={Caelles, Sergi and Maninis, Kevis-Kokitsi and Pont-Tuset, Jordi and Leal-Taix{\'e}, Laura and Cremers, Daniel and Van Gool, Luc},
  booktitle={Proceedings of the IEEE conference on computer vision and pattern recognition},
  pages={221--230},
  year={2017}
}

@inproceedings{oh2019video,
  title={Video object segmentation using space-time memory networks},
  author={Oh, Seoung Wug and Lee, Joon-Young and Xu, Ning and Kim, Seon Joo},
  booktitle={Proceedings of the IEEE/CVF international conference on computer vision},
  pages={9226--9235},
  year={2019}
}

@article{wang2018semi,
  title={Semi-supervised video object segmentation with super-trajectories},
  author={Wang, Wenguan and Shen, Jianbing and Porikli, Fatih and Yang, Ruigang},
  journal={IEEE transactions on pattern analysis and machine intelligence},
  volume={41},
  number={4},
  pages={985--998},
  year={2018},
  publisher={IEEE}
}

@inproceedings{park2021learning,
  title={Learning dynamic network using a reuse gate function in semi-supervised video object segmentation},
  author={Park, Hyojin and Yoo, Jayeon and Jeong, Seohyeong and Venkatesh, Ganesh and Kwak, Nojun},
  booktitle={Proceedings of the IEEE/CVF conference on computer vision and pattern recognition},
  pages={8405--8414},
  year={2021}
}

@inproceedings{bhat2020learning,
  title={Learning what to learn for video object segmentation},
  author={Bhat, Goutam and Lawin, Felix J{\"a}remo and Danelljan, Martin and Robinson, Andreas and Felsberg, Michael and Van Gool, Luc and Timofte, Radu},
  booktitle={European Conference on Computer Vision},
  pages={777--794},
  year={2020},
  organization={Springer}
}

@article{fan2021semi,
  title={Semi-supervised video object segmentation via learning object-aware global-local correspondence},
  author={Fan, Jiaqing and Liu, Bo and Zhang, Kaihua and Liu, Qingshan},
  journal={IEEE Transactions on Circuits and Systems for Video Technology},
  volume={32},
  number={12},
  pages={8153--8164},
  year={2021},
  publisher={IEEE}
}

@article{sun2023munet,
  title={Munet: Motion uncertainty-aware semi-supervised video object segmentation},
  author={Sun, Jiadai and Mao, Yuxin and Dai, Yuchao and Zhong, Yiran and Wang, Jianyuan},
  journal={Pattern Recognition},
  volume={138},
  pages={109399},
  year={2023},
  publisher={Elsevier}
}

@inproceedings{lu2019see,
  title={See more, know more: Unsupervised video object segmentation with co-attention siamese networks},
  author={Lu, Xiankai and Wang, Wenguan and Ma, Chao and Shen, Jianbing and Shao, Ling and Porikli, Fatih},
  booktitle={Proceedings of the IEEE/CVF conference on computer vision and pattern recognition},
  pages={3623--3632},
  year={2019}
}

@inproceedings{hu2018unsupervised,
  title={Unsupervised video object segmentation using motion saliency-guided spatio-temporal propagation},
  author={Hu, Yuan-Ting and Huang, Jia-Bin and Schwing, Alexander G},
  booktitle={Proceedings of the European conference on computer vision (ECCV)},
  pages={786--802},
  year={2018}
}

@inproceedings{wang2019learning,
  title={Learning unsupervised video object segmentation through visual attention},
  author={Wang, Wenguan and Song, Hongmei and Zhao, Shuyang and Shen, Jianbing and Zhao, Sanyuan and Hoi, Steven CH and Ling, Haibin},
  booktitle={Proceedings of the IEEE/CVF conference on computer vision and pattern recognition},
  pages={3064--3074},
  year={2019}
}

@inproceedings{li2018unsupervised,
  title={Unsupervised video object segmentation with motion-based bilateral networks},
  author={Li, Siyang and Seybold, Bryan and Vorobyov, Alexey and Lei, Xuejing and Kuo, C-C Jay},
  booktitle={Proceedings of the European conference on computer vision (ECCV)},
  pages={207--223},
  year={2018}
}

@inproceedings{cheng2021modular,
  title={Modular interactive video object segmentation: Interaction-to-mask, propagation and difference-aware fusion},
  author={Cheng, Ho Kei and Tai, Yu-Wing and Tang, Chi-Keung},
  booktitle={Proceedings of the IEEE/CVF conference on computer vision and pattern recognition},
  pages={5559--5568},
  year={2021}
}

@inproceedings{miao2020memory,
  title={Memory aggregation networks for efficient interactive video object segmentation},
  author={Miao, Jiaxu and Wei, Yunchao and Yang, Yi},
  booktitle={Proceedings of the IEEE/CVF conference on computer vision and pattern recognition},
  pages={10366--10375},
  year={2020}
}

@inproceedings{oh2019fast,
  title={Fast user-guided video object segmentation by interaction-and-propagation networks},
  author={Oh, Seoung Wug and Lee, Joon-Young and Xu, Ning and Kim, Seon Joo},
  booktitle={Proceedings of the IEEE/CVF conference on computer vision and pattern recognition},
  pages={5247--5256},
  year={2019}
}

@inproceedings{heo2020interactive,
  title={Interactive video object segmentation using global and local transfer modules},
  author={Heo, Yuk and Jun Koh, Yeong and Kim, Chang-Su},
  booktitle={European Conference on Computer Vision},
  pages={297--313},
  year={2020},
  organization={Springer}
}

@article{zhao2023fast,
  title={Fast segment anything},
  author={Zhao, Xu and Ding, Wenchao and An, Yongqi and Du, Yinglong and Yu, Tao and Li, Min and Tang, Ming and Wang, Jinqiao},
  journal={arXiv preprint arXiv:2306.12156},
  year={2023}
}

@article{zhang2023faster,
  title={Faster segment anything: Towards lightweight sam for mobile applications},
  author={Zhang, Chaoning and Han, Dongshen and Qiao, Yu and Kim, Jung Uk and Bae, Sung-Ho and Lee, Seungkyu and Hong, Choong Seon},
  journal={arXiv preprint arXiv:2306.14289},
  year={2023}
}

@inproceedings{xiong2024efficientsam,
  title={Efficientsam: Leveraged masked image pretraining for efficient segment anything},
  author={Xiong, Yunyang and Varadarajan, Bala and Wu, Lemeng and Xiang, Xiaoyu and Xiao, Fanyi and Zhu, Chenchen and Dai, Xiaoliang and Wang, Dilin and Sun, Fei and Iandola, Forrest and others},
  booktitle={Proceedings of the IEEE/CVF Conference on Computer Vision and Pattern Recognition},
  pages={16111--16121},
  year={2024}
}

@article{zhou2023edgesam,
  title={Edgesam: Prompt-in-the-loop distillation for on-device deployment of sam},
  author={Zhou, Chong and Li, Xiangtai and Loy, Chen Change and Dai, Bo},
  journal={arXiv preprint arXiv:2312.06660},
  year={2023}
}

@inproceedings{shu2025tinysam,
  title={TinySAM: Pushing the envelope for efficient segment anything model},
  author={Shu, Han and Li, Wenshuo and Tang, Yehui and Zhang, Yiman and Chen, Yihao and Li, Houqiang and Wang, Yunhe and Chen, Xinghao},
  booktitle={Proceedings of the AAAI Conference on Artificial Intelligence},
  volume={39},
  number={19},
  pages={20470--20478},
  year={2025}
}

@article{ma2024segment,
  title={Segment anything in medical images},
  author={Ma, Jun and He, Yuting and Li, Feifei and Han, Lin and You, Chenyu and Wang, Bo},
  journal={Nature Communications},
  volume={15},
  number={1},
  pages={654},
  year={2024},
  publisher={Nature Publishing Group UK London}
}

@article{wu2025medical,
  title={Medical sam adapter: Adapting segment anything model for medical image segmentation},
  author={Wu, Junde and Wang, Ziyue and Hong, Mingxuan and Ji, Wei and Fu, Huazhu and Xu, Yanwu and Xu, Min and Jin, Yueming},
  journal={Medical image analysis},
  volume={102},
  pages={103547},
  year={2025},
  publisher={Elsevier}
}

@article{mazurowski2023segment,
  title={Segment anything model for medical image analysis: an experimental study},
  author={Mazurowski, Maciej A and Dong, Haoyu and Gu, Hanxue and Yang, Jichen and Konz, Nicholas and Zhang, Yixin},
  journal={Medical Image Analysis},
  volume={89},
  pages={102918},
  year={2023},
  publisher={Elsevier}
}

@inproceedings{ding2025sam2long,
  title={Sam2long: Enhancing sam 2 for long video segmentation with a training-free memory tree},
  author={Ding, Shuangrui and Qian, Rui and Dong, Xiaoyi and Zhang, Pan and Zang, Yuhang and Cao, Yuhang and Guo, Yuwei and Lin, Dahua and Wang, Jiaqi},
  booktitle={Proceedings of the IEEE/CVF International Conference on Computer Vision},
  pages={13614--13624},
  year={2025}
}

@inproceedings{videnovic2025distractor,
  title={A distractor-aware memory for visual object tracking with sam2},
  author={Videnovic, Jovana and Lukezic, Alan and Kristan, Matej},
  booktitle={Proceedings of the Computer Vision and Pattern Recognition Conference},
  pages={24255--24264},
  year={2025}
}

}

\clearpage
\setcounter{page}{1}

\maketitlesupplementary

\section{Appendix} \label{sec:appendix}

\begin{figure*}[!b]
    \vspace{0.5cm}
    \centering
    \rotatebox{90}{\includegraphics[width=12cm]{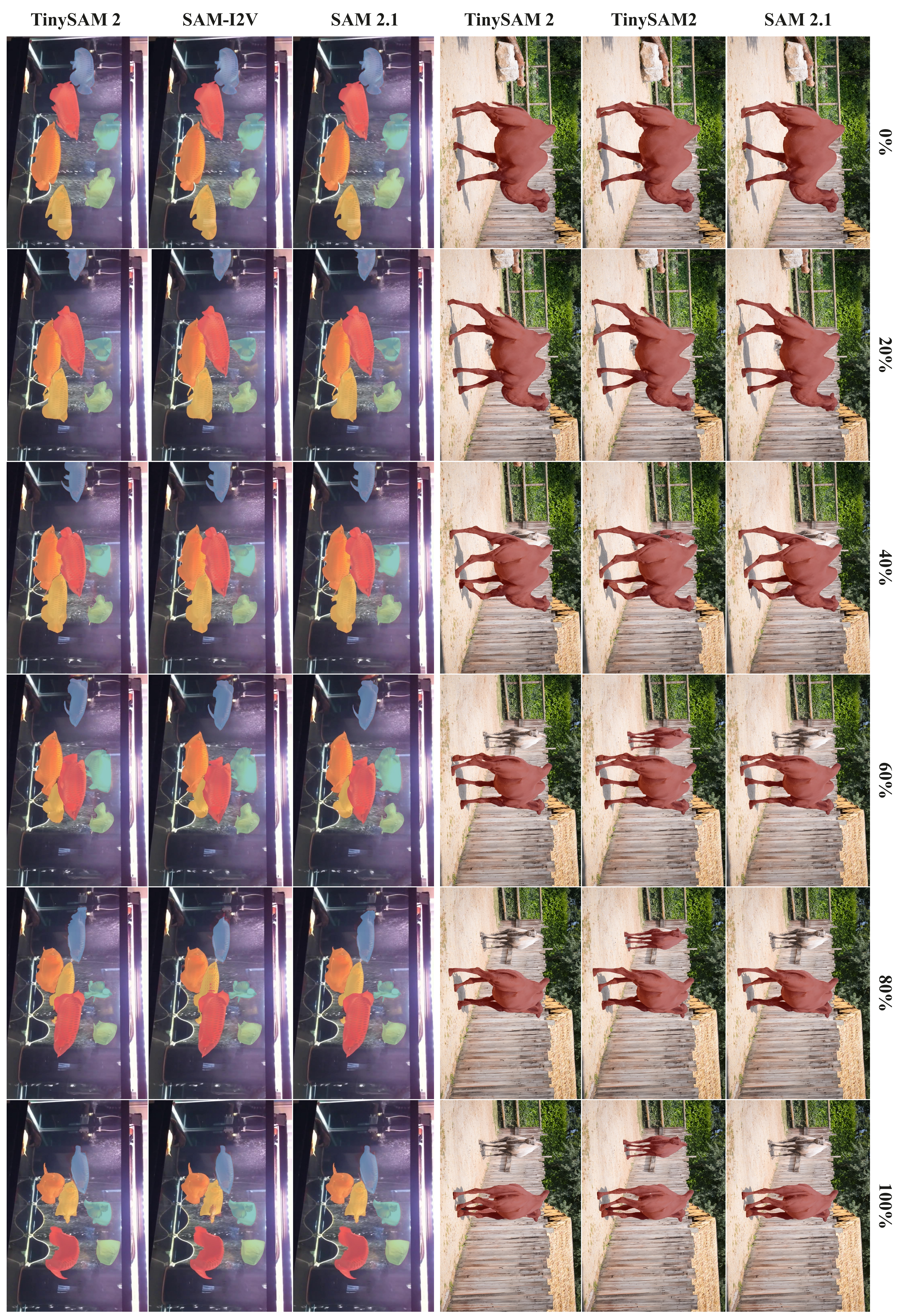}}
    \caption{\textbf{Qualitative comparison of TinySAM 2, SAM 2.1 and SAM-I2V in videos across different scenarios.} We visually demonstrate the results by uniformly sampling frames from complete videos of different scenes. The top example is a camel, the bottom one is an aquarium. In the top example, SAM-I2V incorrectly tracks objects behind the target, whereas TinySAM 2 clearly outperforms SAM-I2V in this aspect.}
    
    \label{fig_c6_app1}
    \vspace{0.5cm}
\end{figure*}

\begin{figure*}[!h]
    
    \centering
    \rotatebox{90}{\includegraphics[width=18.2cm]{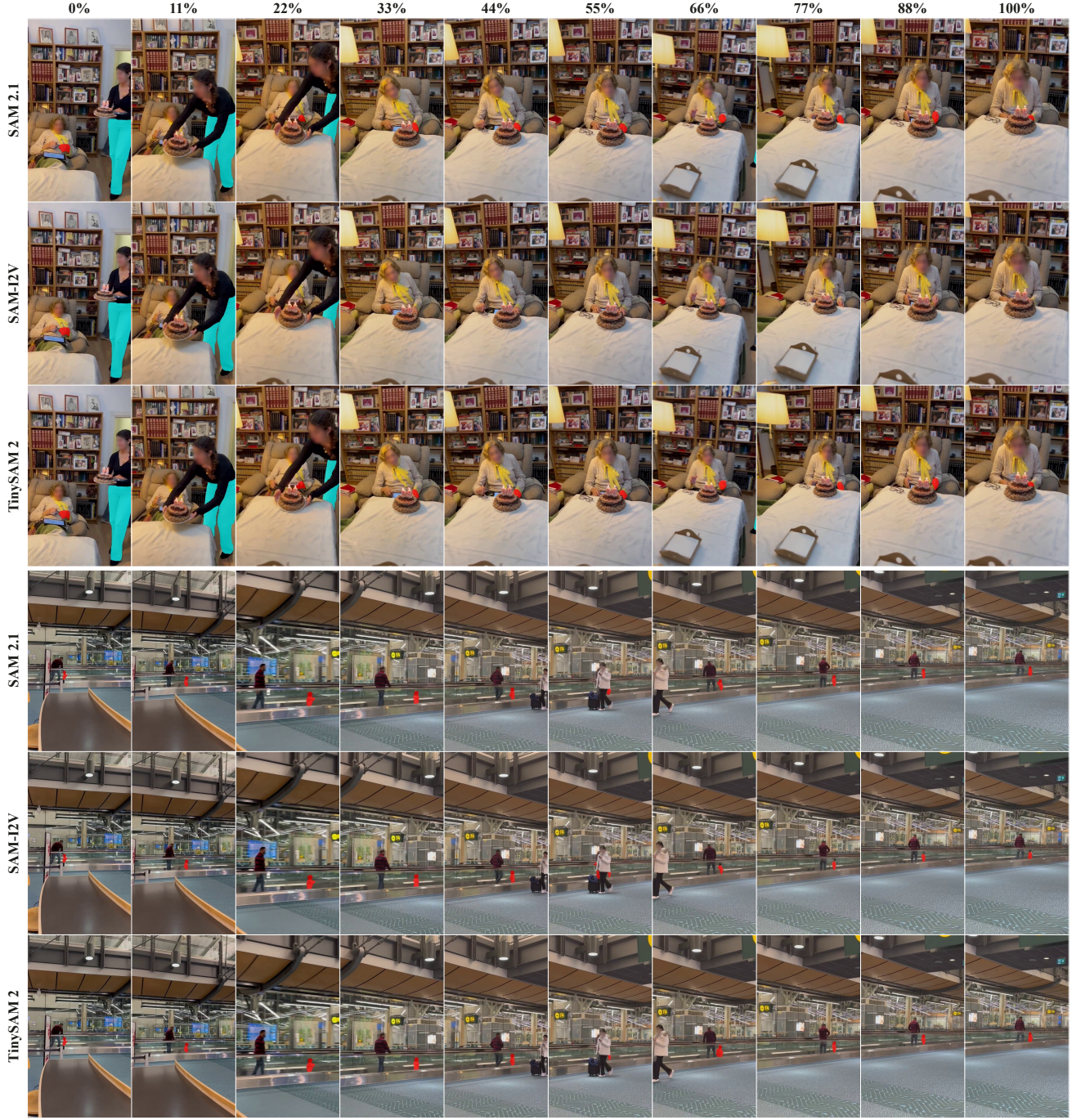}}
    \caption{\textbf{Qualitative comparison of TinySAM 2, SAM 2.1 and SAM-I2V on the SA-V dataset.} The challenge of the SA-V dataset lies in the longer video lengths and the fact that the tracked objects are either partial or relatively small in size. The example on top shows a birthday scene, while the one below depicts an airport. SAM-I2V encounters issues with tracking failures or incomplete tracking when the target is occluded, whereas TinySAM 2 clearly outperforms SAM-I2V in this aspect.}
    \vspace{-.3cm}
    \label{fig_c6_app2}
\end{figure*}

We further demonstrate the visual comparisons of TinySAM 2, SAM 2.1, and SAM-I2V across different videos. 
Figure~\ref{fig_c6_app1} illustrates the ability to accurately segment movements among animals.
Figure~\ref{fig_c6_app2} demonstrates the capability to precisely track small-sized 
or local parts of objects over more than 400 frames of video.
In Figure~\ref{fig_c6_app1}, starting at 40\% of the video, SAM-I2V mistakenly begins tracking a camel that appears behind the tracked camel. In the second example, starting at 80\% of the video, SAM-I2V starts to confuse the red-marked fish with the yellow-marked fish that follows it.
In Figure~\ref{fig_c6_app2} , starting at 11\% of the video, SAM-I2V begins to lose track of the hand that is partially obscured by a cake, and also loses partial tracking of the leg that was previously occluded. In the second example, starting at 55\% of the video, SAM-I2V incorrectly tracks a pedestrian who is causing the occlusion, thus losing track of the child's head in the subsequent frames.
Although there are differences in segmentation results among different methods, our approach consistently shows competitiveness, especially when compared to SAM-I2V in terms of accurately tracking objects. This clearly proves the effectiveness of our method.

\end{document}